\documentclass[letterpaper, 10 pt, conference]{ieeeconf}  

\IEEEoverridecommandlockouts                              

\usepackage{graphicx} 
\usepackage{xcolor}
\usepackage{amsmath}
\usepackage{booktabs}
\usepackage{amssymb}
\usepackage{algorithm}
\usepackage{algpseudocode}
\usepackage{cite}

\title{\vspace{-3em}Learning Compositional Symbolic Task Rules from Demonstrations with Inductive Logic Programming\vspace{-0em}}

\author{Oleh Borys$^{1}$ and Karla Stepanova$^{1}$%
\thanks{$^{1}$ The authors are with the Czech Institute of Informatics, Robotics and Cybernetics,
Czech Technical University in Prague.}%
\thanks{This work was co-funded by the European Union under the project Robotics and Advanced
Industrial Production (reg. no. CZ.02.01.01/00/22\_008/0004590) and by the Junior STAR project
PersonalRobot no.~26-22610M, funded by the Czech Science Foundation.}%
}

\begin{document}

\maketitle
\thispagestyle{empty}
\pagestyle{empty}

\begin{abstract}
Learning from Demonstration~(LfD) should capture not only how a task is executed, but also its high-level task structure that explains the demonstrated behavior. As robots become more autonomous, such task representations must be inspectable, reusable, and human-interpretable. To address this, we study how to represent and learn robotic tasks with inductive logic programming~(ILP) by decomposing a complex task into a series of simpler learning objectives at different abstraction (ontological) levels. The system infers symbolic rules from demonstrations and prior (domain) knowledge, and reuses learned rules when learning higher-level task structure. We evaluate the approach in a synthetic block-assembly scenario and show that the learned abstractions are interpretable and support strong generalization to harder, held-out tasks with unseen objects. These results provide preliminary evidence that decomposed ILP is a feasible approach to task-level LfD.

\end{abstract}

\section{INTRODUCTION}
\label{sec:introduction}
Learning from Demonstration~(LfD) is often framed as learning how to reproduce a task (i.e., \textit{how to act}), but for many robotic problems, this is not enough. The robot should also recover the symbolic high-level task structure underlying the demonstrations (i.e., \textit{what to do}), in the form of abstractions and rules~\cite{ivntr,visualpredicator, embod_abstr, silver2023learning} that are interpretable, easy to inspect, and reusable. These tasks and world concepts can lie at different abstraction (ontological) levels, so the system goes beyond learning statistical patterns from demonstrations and induces the corresponding logic-rich rules at each level.

\begin{figure}[!h]
\includegraphics[width = \linewidth]{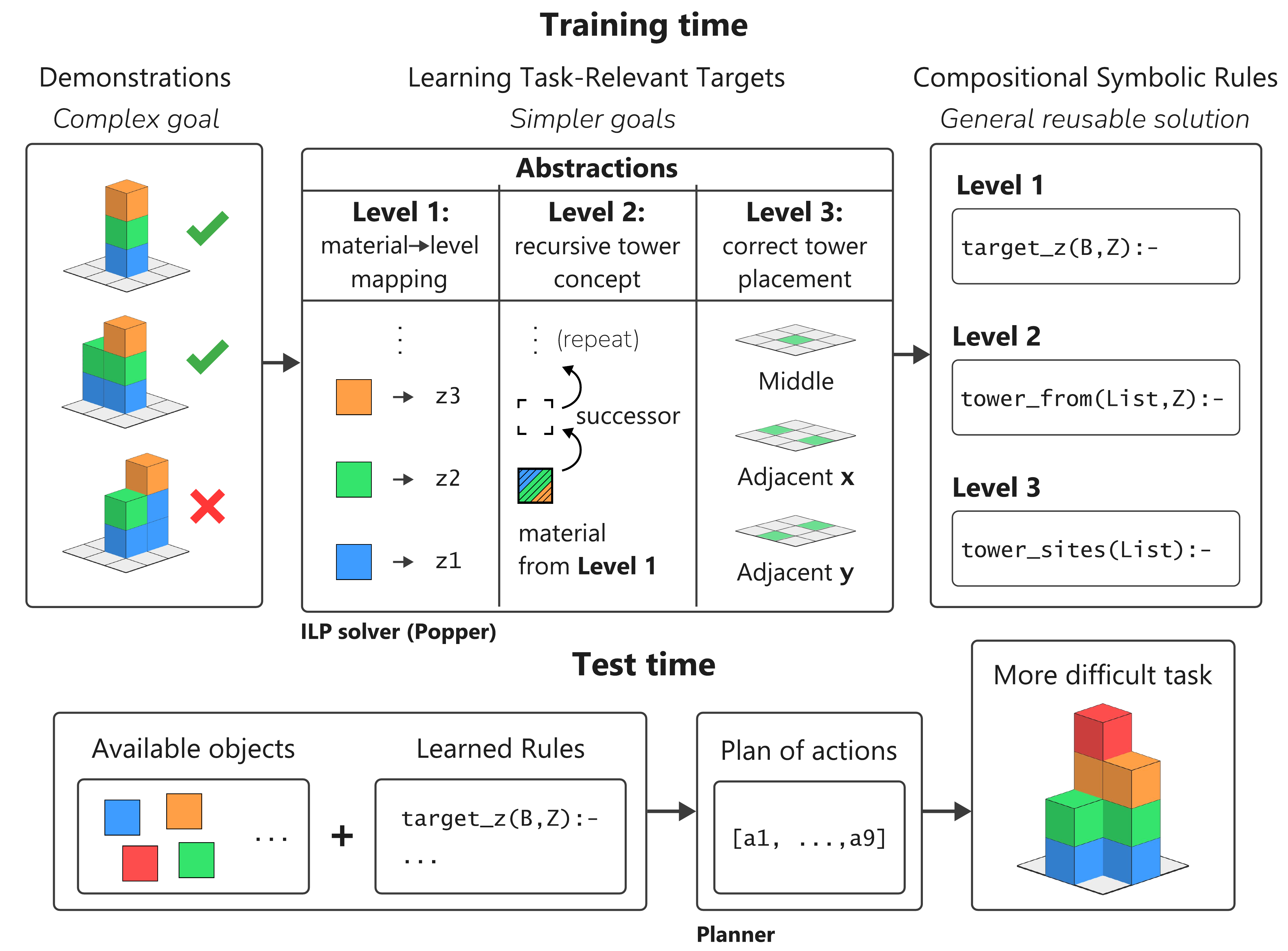}
\vspace{-1.7em}
\caption{Overview of our task-level LfD pipeline.}%
\vspace{-1em}
\label{fig:system}
\end{figure}

Consider, for example, an assembly task where a robot must build several towers from blocks (see Fig.~\ref{fig:system}). Each tower may have a different height, and the block material required at each tower level may vary. Blocks may also have other properties than material that are irrelevant to the task. Although the overall task appears logically complex, it can actually be broken down into simpler tasks: (i) learning which material belongs at each level, (ii) learning the concept of a tower with variable height by reusing the learned material-to-level mapping, and (iii) learning where to place towers. This suggests that task-level LfD can benefit from decomposing the complex task into simpler, interpretable learning goals on different levels of abstraction hierarchy.

In this workshop paper, we study how inductive logic programming~(ILP), specifically Popper~\cite{popper}, can be used to induce interpretable, relational task rule sets from demonstrations and background knowledge. Prior work has applied ILP and ILP-related relational rule learning in robotics to learn action models, task specifications, tool-use constraints, and affordances~\cite{ilp_inductive,ilp_rel_affordance,ilp_robot_engineer,ilp_tool_use}. Still, the area remains mostly underexplored in robotics, particularly in task-level LfD. We argue that ILP is a promising fit for this setting because it naturally casts task learning as inducing compact rules that best explain positive and negative demonstrations under explicit background knowledge and bias.

We decompose a complex task into a sequence of simpler task-relevant learning targets at different levels of abstraction, and then, for each of them, we use Popper~\cite{popper} to induce symbolic rules over the allowed background predicates and add the learned rules back into the background knowledge for reuse in subsequent targets. The result is a planning problem constrained by the union of learned hierarchical rules. At execution time, a planner produces a pick-and-place action sequence that satisfies these constraints.

Existing task-level approaches often rely on latent neural representations, neural rule-learning modules, or stochastic grounding of first-order predicates~\cite{ivntr,visualpredicator}, providing only partial or post-hoc interpretability for logically complex concepts. For example, VisualPredicator~\cite{visualpredicator} uses VLM-generated neuro-symbolic predicates, but VLM-based evaluation can be unreliable for logically complex concepts such as \texttt{\small{jug\_filled(J)}}. In contrast, we express such concepts compositionally as rules over simpler predicates, so only simple atoms need to be evaluated, and their confidence can be propagated through logical inference. Our system also learns multi-level abstractions that build on one another, and the preliminary experiments show reduced training and inference costs compared with these methods. 

Our approach is primarily aimed at manipulation problems that can be easily and explicitly defined by human rules, e.g., sorting, tidying a table, and assembly tasks. The main contributions of this paper are:
\begin{enumerate}
    \item We present an ILP-based framework for task-level LfD that learns and reuses first-order logic compositional rules across abstraction levels.
    \item We provide a qualitative and quantitative evaluation in a synthetic block-assembly domain, showing that the learned rules are interpretable, support planning, and generalize to harder test tasks with unseen objects and placement locations.
\end{enumerate}

\section{RELATED WORK}
\label{sec:related_works}
\subsection{Learning abstractions, planning, and reasoning}
Recent work in robotics has explored integrating neural perception with symbolic reasoning to obtain task-relevant abstractions for planning and control~\cite{ivntr,visualpredicator,clier}. IVNTR~\cite{ivntr} combines bilevel planning with bilevel learning - it makes the symbolic part invent predicates and the neural part learn their meaning, guiding each other. The authors explicitly state that it is hard to interpret the physical meaning of the predicates because they are neural networks. 
Another approach, VisualPredicator~\cite{visualpredicator}, introduces Neuro-Symbolic Predicates (NSPs), represented as Python programs that query vision-language models (VLMs) and are composed into symbolic operators for planning. The approach remains interpretable and enables very flexible predicate invention. However, VLM-based evaluation can be unreliable for resolving logically complex concepts in one step. Instead, we propose to decompose them into rules over simpler predicates, which could be evaluated much more easily. The resulting confidences for these atoms can then be propagated through logical inference to estimate the confidence of the higher-level predicate. This approach still leverages VLM strengths for perception, but improves reliability and interpretability.

\subsection{Inductive logic programming (Popper)}
Inductive logic programming (ILP) has advantages over statistical learning methods: expressiveness of first-order logic, learning from a few examples thanks to the strong inductive bias by background knowledge, easy rules use and transfer thanks to its symbolic nature~\cite{ilp_intro}. Popper~\cite{popper} is the state-of-the-art ILP approach that implements the concept of learning from failures (LFF). Comparative studies~\cite{ilp_and_popper} highlight its strong empirical performance and search efficiency. Its support for recursive and optimal rule learning makes it particularly suitable for our use case. While ILP has been explored for some symbolic robotics problems, its use as the core mechanism for task-level learning from demonstration remains comparatively underexplored; in contrast, we use Popper~\cite{popper} to learn an ordered sequence of reusable task rules from demonstrations and background knowledge, and then use these rules to constrain planning.

\section{PROBLEM FORMULATION}
\label{sec:problem_formulation}
We study task-level LfD in a discrete manipulation domain (e.g., sorting or block assembly) as the problem of recovering an explicit \emph{task knowledge} from observed executions. We focus on the symbolic decision-making layer: the robot operates with parameterized pick-and-place actions, and we do not model low-level motion planning.

A task is a tuple \mbox{$T = \langle \mathcal{O}, x_0, X_g \rangle$}, sampled from a distribution \mbox{$T \sim \mathcal{T}$}, where \mbox{$\mathcal{O}=\{o_1,\dots,o_N\}$} is the set of objects, \mbox{$x_0 \in X$} is an initial state, and \mbox{$X_g\subseteq X$} is the set of goal-satisfying states. Each state is a discrete encoding of object positions. We assume a given parameterized action schema set \mbox{$\mathcal{A}$} and a (deterministic) task-level transition function \mbox{$f: X \times \mathcal{A} \rightarrow X$} with \mbox{$x_{t+1} = f(x_t,a_t)$}. At execution time, the task-level objective is to output a finite action sequence \mbox{$\pi=[a_0,\dots, a_{H-1}]$}, \mbox{$a_t\in\mathcal{A}$}, such that the induced trajectory \mbox{$[x_0,\dots,x_H]$} reaches a goal-satisfying state, i.e., \mbox{$x_H\in X_g$}. During training, we observe an offline dataset of demonstrations \mbox{$\mathcal{D}=\mathcal{D}^+\cup\mathcal{D}^-$} consisting of labeled finite state sequences \mbox{$d=[x_0,\dots,x_H]$}.

Let $\Psi$ be a predicate vocabulary describing object properties and relations. A \emph{ground atom} is a predicate applied to constants, e.g., \mbox{\texttt{\small{material(B,wood)}}}. We use Prolog-style definite clauses (Horn clauses), written as \mbox{\texttt{\small{h(X):- b1(Y1),\dots,bm(Ym)}}}, where the body is a conjunction of atoms.
We represent prior knowledge as a background program~$B$ (a set of facts and rules) containing object ontology and environment definitions such as grid topology.

To connect low-level states to logic, a grounding mechanism for \emph{atomic} predicates \mbox{$\Psi_0\subseteq\Psi$} returns a set $\Gamma(x)$ of true ground atoms with \mbox{$\Gamma(x) \subseteq \{\psi(\bar{c}) \mid \psi\in\Psi_0,\; \bar{c}\in\mathcal{C}^{\mathrm{arity}(\psi)}\}$} for each state \mbox{$x\in X$}, where $\mathcal{C}$ denotes the available constant symbols and $\bar{c}$ is a tuple of constants. We define the resulting abstract theory as \mbox{$\Gamma_B(x) \triangleq B \cup \Gamma(x)$} and treat a ground atom $a$ as true in $x$ whenever \mbox{$\Gamma_B(x)\models a$}, allowing higher-level predicates to be derived compositionally.

\section{METHODOLOGY}
\label{sec:methodology}
In this work, we assume that (i) the predicate vocabulary ~$\Psi$, target list ~$\Psi_{\text{learn}}$, and the corresponding ILP biases are user-specified, (ii) the grounding function~$\Gamma(\cdot)$ is given, and (iii) all features are discrete or discretized. We use inductive logic programming~(ILP), specifically Popper~\cite{popper}, to induce symbolic task rules from demonstrations and background knowledge. Our method is summarized in Fig.~\ref{fig:system}.

A background program~$B$ (facts and rules) is encoded in Prolog~\cite{prolog}, together with a user-provided language bias (allowed body predicates, types/modes, recursion settings, and syntactic limits). 
Learning a unified task-level program would require broad background knowledge and flexible bias for Popper~\cite{popper}, quickly blowing up the hypothesis search space and making the learning difficult, if not impossible. Instead, we structure learning as an ordered list of task-relevant targets \mbox{$\Psi_{\text{learn}}=[\tau_1,\dots,\tau_K]$}. Each target~$\tau$ captures a particular \emph{viewpoint} on the task (e.g., material-to-level constraints, recursively defined tower structure, or placement rules), and later targets may be defined in terms of previously learned ones. For each target predicate \mbox{$\tau\in\Psi_{\text{learn}}$}, we ground each demonstration \mbox{$x\in\mathcal{D}$} into an abstract state \mbox{$\Gamma_B(x)=B\cup\Gamma(x)$} and extract labeled examples $E_\tau^+$ and $E_\tau^-$ for~$\tau$. Popper~\cite{popper} then induces a hypothesis~$H_\tau$ (a set of definite clauses with head predicate~$\tau$), as defined in Eq.~\ref{eq:popper}.

\begingroup
\setlength{\abovedisplayskip}{-0.8ex}
\setlength{\abovedisplayshortskip}{-0.8ex}
\begin{equation} \label{eq:popper}
B \cup H_\tau \models e \;\;\; \forall e\in E_\tau^+,\qquad B \cup H_\tau \not\models e \;\;\; \forall e\in E_\tau^-.
\end{equation}
  
After learning each~$\tau$, we augment the background knowledge \mbox{$B\leftarrow B\cup H_\tau$} to enable reuse of learned rules, yielding a compact multi-level abstraction hierarchy. The resulting overall hypothesis is \mbox{$H=\bigcup_{\tau\in\Psi_{\text{learn}}}H_\tau$}.

At test time, we run a planner written in Prolog~\cite{prolog} that, given the available objects, enumerates candidate sequences and uses the learned rules~$H$ in applicability/transition checks to prune them. It returns the first goal-reaching plan and compiles it into pick-and-place actions. The planner is just a consumer of $H$ and can be replaced.

\section{EXPERIMENTAL EVALUATION}
\label{sec:experimental_evaluation}

\subsection{Experimental setup}

We evaluate our approach in a synthetic block-assembly domain in which a robot builds vertical towers in a \mbox{$3 \times3$}~grid world (see Fig.~\ref{fig:system}). Towers have heights of \mbox{$2$--$4$}~blocks and consist of blocks with properties such as material, color, and shape. All experiments were conducted on a single AMD Ryzen~9~7900~CPU. For each task, inducing rules for all targets and planning at test time took a few seconds in total.

\subsubsection{Train and test tasks} 
The evaluation dataset consists of training demonstrations and held-out test tasks. Each training task contains 12 blocks: 3 stone, 3 brick, 2 glass, and 4 distractors with color and shape but no material label, sufficient to build one tower of height~2 and two towers of height~3. Demonstrations include positive examples of valid constructions and negative examples of invalid structures or placements, see Fig.\ref{fig:system}. Each test task also contains 12 blocks, but includes 1 wood block and only 3 distractors. The test goal is harder: the robot must build towers of heights~2, 3, and 4, and the placement sites used during training are blocked, while previously blocked sites are enabled. Solving these held-out tasks, therefore, requires generalization both to taller towers and to unseen placements.

\subsubsection{Dataset generation} 
Our dataset is generated from a ground-truth rule set~$H_{GT}$. We repeat the experiment $10$~times with different random objects and demo stratifications. In each repetition, we randomly assign $60$~blocks, each with a unique object identifier (\mbox{$4$ material $\times$} \mbox{$5$ color $\times$} \mbox{$3$ shape} combinations) across $4$~training tasks and $1$~test task, with disjoint train/test identities. This yields \mbox{$4 \times$} \mbox{$1 \times 10 = 40$} train--test evaluations. We generate negative demonstrations by permuting objects (e.g., a stone block cannot be on top of a glass block), showing incorrect placement sites, and adding \emph{distraction objects} that must not be used in a correct tower. The demonstration data are created by stratified subsampling: all positive examples are retained, while only a seed-controlled fraction of negative examples is sampled (separate sampling rates and stable per-target offsets).

\subsubsection{Dataset - rule labeling}
We run the planner with~$H_{GT}$ on each task instance to generate labeled demonstrations. For readability, we describe the rules in plain language: ``Build one tower at the anchor location (grid center), and the others at allowed cardinal neighbor sites. Each tower is a recursive list of successor blocks and follows the material order: stone at~\texttt{\small{z1}}, brick at~\texttt{\small{z2}}, glass at~\texttt{\small{z3}}, and wood at~\texttt{\small{z4}}.'' The high-level objective is hand-coded; learning focuses on the structural and placement rules.

\subsection{Learning goal} 
The learner’s goal is to induce three predicates \mbox{$\tau \in \Psi_{learn}$} that capture task-relevant structure: \texttt{\small{target\_z/2}}, which maps object properties to the vertical level, \texttt{\small{tower\_from/2}}, which specifies which set of objects constitutes a tower, and \texttt{\small{tower\_site/1}}, which characterizes admissible tower locations given constraints, see Fig.\ref{fig:system}. Our initial background knowledge~$B$ encodes: object/property facts (e.g., \texttt{\small{block/1, color/2}}), constants as unary predicates for materials and height levels (e.g., \texttt{\small{stone/1}}, \texttt{\small{z1/1}}), and relational and spatial predicates (\texttt{\small{succ\_z/2}}, \texttt{\small{adj\_h(S1, S2)}}). 

To \emph{test modularity}, the training tasks intentionally omit examples of placing wooden blocks at \texttt{\small{z4}}; instead, the corresponding \mbox{$\texttt{\small{wood}}\mapsto\texttt{\small{z4}}$} mapping is injected directly into~$B$. This could be prior domain knowledge, or knowledge obtained from another source, e.g., a human natural-language instruction translated into deterministic logic. Importantly, this injected mapping alone does not yield height-4 generalization: solving the test tasks also requires learning a sufficiently general (recursive) \texttt{\small{tower\_from/2}} definition.

For each target \mbox{$\tau \in \Psi_{learn}$}, we provide Popper~\cite{popper} with examples, a target-specific language bias, and the current background program~$B$. For \texttt{\small{target\_z/2}}, the bias permits object-property predicates (e.g., \texttt{\small{shape/2}}) and their unary versions. For \texttt{\small{tower\_from/2}}, it enables \texttt{\small{succ\_z/2}}, the learned \texttt{\small{target\_z/2}}, and additional \texttt{\small{head/2}}, \texttt{\small{tail/2}}, \texttt{\small{empty/1}} to support lists (also, recursion is allowed). Finally, for \texttt{\small{tower\_site/1}}, it enables placement predicates such as \texttt{\small{goal\_anchor/1, adj\_h/2}} (plus a distraction predicate \texttt{\small{diagonal\_adj/2}}), and \texttt{\small{head/2}}, \texttt{\small{tail/2}}, \texttt{\small{empty/1}}. 
After learning each target~$\tau$, we add the learned clauses back into the background knowledge (\mbox{$B \gets B \cup H_{\tau}$}) before learning the next one. The final rule set \mbox{$H=\bigcup_{\tau\in\Psi_{\text{learn}}}H_\tau$} is then passed to the planner to produce a pick-and-place action sequence.

\label{sec:quantitative_res}
\begin{figure}[t]
  \centering
  \includegraphics[width=\columnwidth]{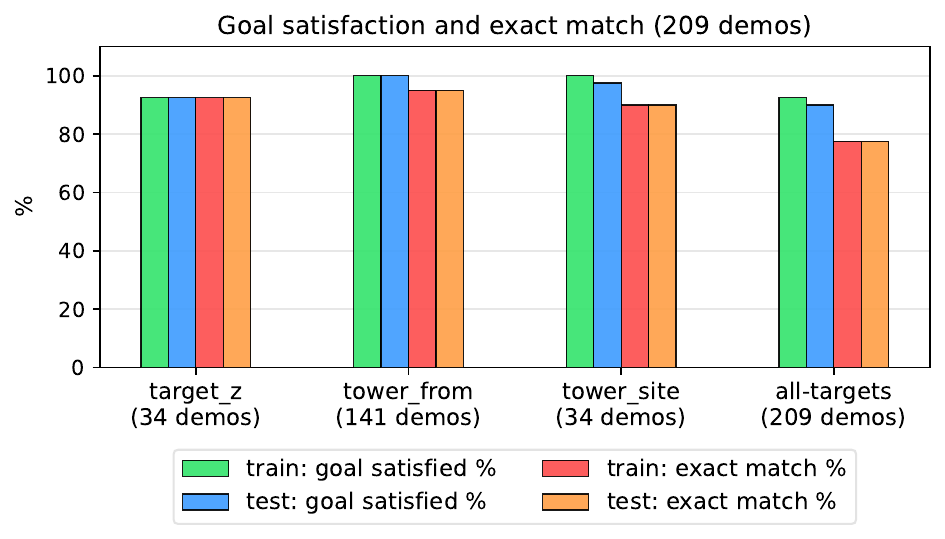}
  \vspace{-2em}
  \caption{Per-target and all-targets learning outcomes, averaged over $10$ repetitions ($209$ demonstrations for each task).}
  \vspace{-1em}
  \label{fig:results}
\end{figure}

\subsection{Quantitative evaluation}

Fig.~\ref{fig:results} reports quantitative per-target and all-targets results, averaged over $10$ repetitions (each task using $209$~demonstrations in total). We report, across training and test tasks, (i) whether the planning goal is satisfied and (ii) whether the learned rules are logically correct. To evaluate a single predicate, we combine its learned rules with ground-truth rules for the remaining targets and compare against the full ground-truth program. To evaluate the all-targets setting, we plan with the full learned program, i.e., the union of the learned clauses for all targets. When evaluating \texttt{\small{target\_z}} predicate specifically, we also add the \texttt{\small{wood}}$ \ \mapsto  \ $\texttt{\small{z4}} mapping rule to the set of learned rules to make the comparison fair. As shown in Fig.~\ref{fig:results}, goal satisfaction can be higher than an exact logical match accuracy, i.e., approximate rules may still be sufficient to solve the task.
 
Nevertheless, we found that the system requires~$40$, $166$, and $37$~demonstrations, on average, for \texttt{\small{target\_z/2}}, \texttt{\small{tower\_from/2}}, and \texttt{\small{tower\_site/1}}, respectively, to achieve both 100\% goal satisfaction and logical rule match for each target and~$243$ demonstrations in total for the all-targets system. We also tried manually selecting demonstrations to pack more constraints into fewer examples. In this setting, we were able to achieve 100\% goal satisfaction and logical rule match with~$13$, $50$, and $14$~demonstrations for each \mbox{$\tau \in \Psi_{learn}$} respectively, resulting in $77$ demonstrations in total. This suggests the system depends strongly on demonstration quality, and better selection strategies could likely reduce the number of required examples.

It is also worth noting that the effective number of unique demonstrations can be much lower than the per-target counts, because the same demonstration can contribute examples to multiple targets. For example, a single demonstration of building a tower, e.g., at location \texttt{\small{s22}}, yields (i) material-to-level supervision for \texttt{\small{target\_z/2}}, (ii) list-structure supervision for \texttt{\small{tower\_from/2}}, and (iii) a valid-location example for \texttt{\small{tower\_site/1}}. When we report the number of demonstrations, we mean the number of positive and negative examples in Popper~\cite{popper} \texttt{\small{exs.pl}} files.

\subsection{Qualitative results}
\label{sec:qualitative_res}

In this section, we show that the learned clauses are compact, interpretable, and applicable. For \texttt{\small{target\_z/2}}, our system learns rules of the form shown in the Clause~\ref{cl:targetz-stone}:

\begingroup
\setlength{\abovedisplayskip}{-2.3ex}
\setlength{\abovedisplayshortskip}{-2.3ex}
\setlength{\belowdisplayskip}{0.5ex}
\setlength{\belowdisplayshortskip}{1.5ex}
\begin{flalign}
\raisetag{0.2\baselineskip}
&\text{\footnotesize\texttt{target\_z(B,Z):- }} \text{\footnotesize\texttt{material(B,M), stone(M), z1(Z).}} \label{cl:targetz-stone}
\end{flalign}
\endgroup

Here \texttt{\small{B}} is a block identifier and \texttt{\small{Z}} is a discrete height level, and the clause states that stone blocks are assigned to the level \texttt{\small{z1}}. 
For \texttt{\small{tower\_from/2}}, it learns a recursive definition, a base case for singleton lists (Clause~\ref{cl:towerfrom-base}), and a recursive step (Clause~\ref{cl:towerfrom-rec}) that advances through successor height levels \texttt{\small{ZNext}}: 

\begingroup
\setlength{\abovedisplayskip}{-2.1ex}
\setlength{\abovedisplayshortskip}{-2.2ex}
\setlength{\belowdisplayskip}{0ex}
\setlength{\belowdisplayshortskip}{1ex}
\setlength{\jot}{0pt} 
\begin{flalign}
\raisetag{0.2\baselineskip}
&\text{\small\texttt{\footnotesize{tower\_from(List,Z):-}}} && \nonumber\\
&\qquad\text{\footnotesize\texttt{head(List,LHead), target\_z(LHead,Z),}} && \nonumber\\ 
&\qquad\text{\footnotesize\texttt{tail(List,LRest), empty(LRest).}} \label{cl:towerfrom-base} &&
\end{flalign}
\endgroup

\begingroup
\setlength{\abovedisplayskip}{-2.3ex}
\setlength{\abovedisplayshortskip}{-2.3ex}
\setlength{\belowdisplayskip}{0.5ex}
\setlength{\belowdisplayshortskip}{0.5ex}
\setlength{\jot}{0pt} 
\begin{flalign}
\raisetag{0.5\baselineskip}
&\text{\small\texttt{\footnotesize{tower\_from(List,Z):-}}} && \nonumber\\
&\qquad\text{\footnotesize\texttt{head(List,LHead), target\_z(LHead,Z),}} && \nonumber\\ 
&\qquad\text{\small\texttt{tail(List,LRest), succ\_z(Z,ZNext),}} && \nonumber\\
&\qquad\text{\footnotesize\texttt{tower\_from(LRest,ZNext).}} \label{cl:towerfrom-rec} &&
\end{flalign}
\endgroup

This recursive structure enables generalization to taller unseen towers as long as \texttt{\small{target\_z/2}} is defined for the required levels. For \texttt{\small{tower\_site/1}}, our system learns rules that characterize admissible site lists relative to the anchor location and its horizontal or vertical neighbors (using list-related predicates and \texttt{\small{goal\_anchor/1, adj\_h/2, adj\_v/2}}), rather than memorizing specific coordinates. This is important for transfer, because the valid test sites differ from those used during training. 

Overall, the qualitative results show that the system recovers a reusable symbolic structure that is directly inspectable and supports out-of-distribution generalization in the held-out tasks (Fig.~\ref{fig:results}).

\section{CONCLUSION AND DISCUSSION}
\label{sec:conclusion_discussion}
We presented an inductive logic programming~(ILP) framework (that uses Popper~\cite{popper}) for task-level Learning from Demonstration~(LfD) that induces human-interpretable first-order logic (FOL) rules from demonstrations and background knowledge. Our approach decomposes a complex manipulation task into a sequence of simpler learning targets across abstraction levels, reusing learned rules from previous stages as background knowledge for later stages. We evaluate the method in a synthetic block-assembly domain, ~Sec.\ref{sec:experimental_evaluation}, and show that the induced rules are interpretable and support OOD generalization. The results provide preliminary evidence that this decomposition makes ILP feasible in practice. Although this paper focuses on the symbolic core and synthetic evaluation, the framework provides a basis for future integration of perception modules and probabilistic reasoning to handle noisy real-world inputs.

A primary limitation is the system's lack of autonomy. While Sec.~\ref{sec:experimental_evaluation} demonstrates how ILP could be used to induce rules for the task, the current framework relies on manual engineering: target predicates, demonstration encodings, and biases are assumed to be given, as required by Popper~\cite{popper}. While restrictive biases enable ILP to learn from small amounts of data, they also require careful tuning. Also, if the learning problem is too big (e.g., extensive background knowledge and bias, enabled recursion or predicate invention), Popper~\cite{popper} may fail to find a solution in a reasonable time, or at all, as searching over FOL hypotheses is a combinatorial problem. Additionally, the current system assumes a discrete state representation because Popper cannot work directly with raw continuous features.

Looking forward, we aim to increase the autonomy of the framework through automated predicate invention and bias generation~\cite{pred_inv,abstr_learn}, potentially complemented by natural language guidance. To handle the combinatorial explosion and ensure that the system captures only goal-relevant effects, we want to implement prioritization via semantic relevance to the task during hypothesis generation. Also, we plan to move to probabilistic logic from FOL, connect perception, and enable predicate evaluation, e.g., by adding Scallop~\cite{scallop2023,scallop2021}, as suggested in~\cite{propper}. Additional extensions include incremental learning from newly observed demonstrations and incorporating active learning from human feedback.

\nocite{*}

\bibliographystyle{IEEEtran}

\bibliography{bibliography_iros2026}  

\end{document}